\documentclass[letterpaper, 10 pt, conference]{ieeeconf}  

\IEEEoverridecommandlockouts                              

\usepackage{multirow}
\usepackage{graphics} 
\usepackage{epsfig} 
\usepackage{mathptmx} 
\usepackage{times} 
\usepackage{amsmath} 
\usepackage{amssymb}  
\usepackage[utf8]{inputenc}
\usepackage{subcaption}
\usepackage{pdfpages}
\usepackage{paralist}
\usepackage[font=small,labelfont=bf]{caption}
\usepackage{array}
\usepackage{booktabs}
\usepackage{lipsum}
\usepackage{caption}
\usepackage{siunitx}
\usepackage[backend=bibtex,style=ieee]{biblatex}
\usepackage{filecontents}
\usepackage{hyperref}

\title{\LARGE \bf
A Deep Learning Approach for Robust Corridor Following
}

\author{Vishnu Sashank Dorbala$^{1}$, A.H. Abdul Hafez$^{2}$ and C.V. Jawahar$^{1}$%
\thanks{}%
\thanks{$^{1}$ The authors are associated with the International Institute of Information Technology, Hyderabad, India
        {\tt\small vdorbala@gmail.com, jawahar@iiit.ac.in}}%
\thanks{$^{2}$ A.H. Abdul Hafez is with Hasan Kalyoncu University, Gaziantep, Turkey  {\tt\small abdul.hafez@hku.edu.tr}}%
}
\newcolumntype{M}[1]{>{\centering\arraybackslash}m{#1}}
\newcommand{\italize}[1]{\textit{#1}}

\newtoggle{draftbib}

\toggletrue{draftbib}

\iftoggle{draftbib}
  {
   \AtEveryBibitem{%
     \clearname{editor}%
     \clearfield{subtitle}%
     \clearfield{maintitle}%
     \clearfield{mainsubtitle}%
     \clearfield{url}%
     \clearfield{ISSN}%
     \clearfield{doi}}}
  {}
  
\begin{document}
\maketitle
\thispagestyle{empty}
\pagestyle{empty}

\begin{abstract}
For an autonomous corridor following task where the environment is continuously changing, several forms of environmental noise prevent an automated feature extraction procedure from performing reliably.
Moreover, in cases where pre-defined features are absent from the captured data, a well defined control signal for performing the servoing task fails to get produced. 
In order to overcome these drawbacks, we present in this work, using a convolutional neural network (CNN) to directly estimate the required control signal from an image, encompassing feature extraction and control law computation into one single end-to-end framework. In particular, we study the task of autonomous corridor following using a CNN and present clear advantages in cases where a traditional method used for performing the same task fails to give a reliable outcome. We evaluate the performance of our method on this task on a Wheelchair Platform developed at our institute for this purpose.
\end{abstract}

\section{INTRODUCTION}
The task of autonomous corridor following has been well discussed in the past \cite{rev1,rev2,rev3,rev4,uavservo}, especially on smart wheelchair platforms. Several classical works achieving this \cite{following1,following2,servoing2,servoing2_expanded,KAK_navigation,Omnidirectional} use vision based algorithms. They extract selected features from a captured image, and pass them to a control law that computes a corrective velocity signal for adjusting the position of the robot on the corridor. There also exist other approaches that use different sensors  \cite{lidar,sonar,radar}, and follow a similar procedure. 
In all of these works, there is an inherent reliance on a robust feature extraction and tracking step to provide reliable features to the control law. As such, the behaviour of the robot becomes undefined when the system fails to provide these features reliably.

In traditional visual servoing (TVS) approaches, when selected features do not appear in the captured image, or when the extracted features are grossly inaccurate, the control law fails to produce a reliable velocity for servoing the robot along the corridor. 
We can infer from this that 
for a TVS process to take place reliably, three major factors need to be accounted for to a good degree of accuracy.
\begin{enumerate}
    \item Quality image features need to be selected for servoing.
    \item They need to be available in the environment.
    \item A robust algorithm is needed for tracking and extracting these features.
\end{enumerate}
  \begin{figure}[t]
      \centering
    \includegraphics[width=\linewidth]{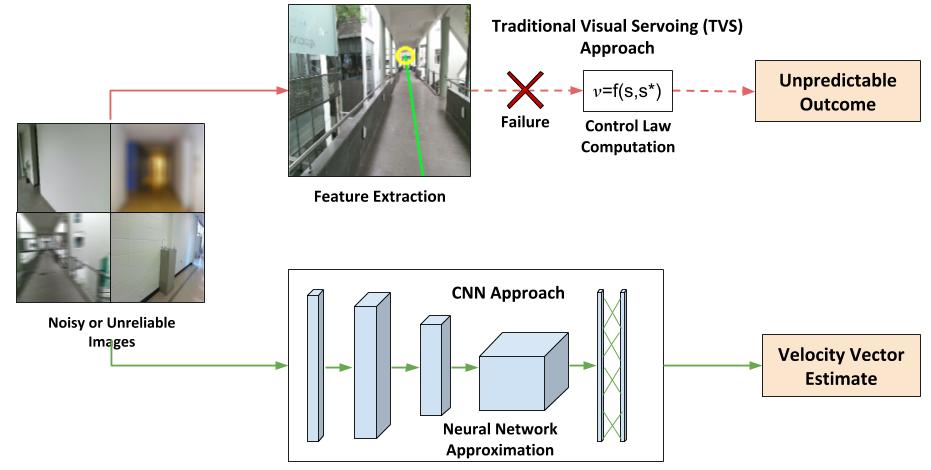}
      \caption[font=small]{Overview of the proposed \italize{CNN Approach} compared to the \italize{ Traditional Visual Servoing (TVS)} used for autonomous corridor following. While the TVS approach performs well on clean images, it often fails to give a reliable output on noisy or occluded images taken from dynamically changing environments. The dotted line represents an unreliable step in the process. In such cases, a neural network can be used to estimate a desirable output.
    }
    \label{fig: Overview1}
    \vspace{-0.7cm}
   \end{figure}


The works presented in \cite{following1,following2,servoing2_expanded} are examples of TVS works. In \cite{following1}, autonomous corridor following is performed on a wheelchair following a TVS approach that
uses vanishing point features from a corridor image for devising a control signal. The work in \cite{following2} extends this to doorway traversal and presents it comprehensively. A similar approach for mobile robots is also proposed in \cite{servoing2} and  extended further in \cite{servoing2_expanded}. Steps for feature extraction which are suggested in these works however do not account for images taken in dynamically changing environments, or for various noise types in the captured images. This hinders with the practical capability of the robot, as motion noise and occlusions are a common occurrence in the environment. Moreover, in cases where the required features cannot be estimated from the image, the outcome of the control law is undefined as it may encounter mathematical singularities. In order to overcome all of these challenges, we propose using a convolutional neural network for approximating a velocity vector output for corridor following directly from a camera image. 
Figure \ref{fig: Overview1} provides an overview of the proposal to solve the problem 
as a robust alternative to traditional visual servoing.

Although there are a few works in the literature that use deep neural networks for visual servoing such as \cite{Madhav_paper,Deep_Learning_Chaumette}, our proposal (Also see \cite{siupaper}) differs from them as we combine both the feature extraction and control signal computation in one stage, while 
they primarily focus on approximating the feature extraction stage. In \cite{Madhav_paper} the authors finetune FlowNet \cite{Flownet1} to estimate the relative angular and translational pose differences between the desired image and the current image. Similarly in \cite{Deep_Learning_Chaumette}, AlexNet has been used to approximate a relative pose between the current image and a reference image. In both these papers, the approximated relative pose is fed into a control law that computes a velocity vector for servoing.
The approach that we present in this work combines the feature extraction and control law computation stages into one framework to directly predict a velocity vector, given an image.

In addition to this, in \cite{QLearning}, a Q learning based approach for visual servoing has been described for performing a target following task using a drone in a simulated environment. They demonstrate the efficacy in using deep features for robust servoing in noisy and occluded environments which further reinforces our usage of deep learning for this visual servoing task.

Our paper makes the following two contributions: i) we introduce a novel CNN based approach for performing an image based visual servoing task of autonomous corridor navigation, and ii) we present a robust comparison showcasing the advantage of our CNN approach against some critical drawbacks of the traditional approach described in \cite{following1}. We carry out a rigorous analytical and practical analysis here to make our case for supporting this claim.

The paper has been organized as follows. Section \ref{sec:2} describes fundamental TVS concepts used for autonomous corridor following, and provides details of our CNN based approach. In Section \ref{Analysis}, we describe the methods we use for robust analysis of our CNN approach where TVS-based approaches fail to perform well. In \ref{Results}, we showcase the results of our experimentation both statistically and practically on a Wheelchair Platform developed at our institute. Finally, we present the advantages of our method by evaluating it on fail cases of the traditional approach.




\section{CNN-Based Autonomous Corridor following}
\label{sec:2}

The basic modelling and control concepts of using TVS in a corridor following task is presented in this section. After that we discuss our CNN-based design along training and data preparation issues.
\subsection{TVS Modelling and Velocity Estimation}
\label{Velocityestimateclassical}
The wheelchair is assumed to be a four wheeled robot with two passive castor wheels in front and two actuated wheels in the rear. It thus behaves as a non-holonomic system constrained by two degrees of freedom.
In order to servo this system along the corridor, a translational velocity $\nu$ and an angular velocity $\omega$ that describe its motion need to be computed.
As our purpose for autonomous corridor following is for assisting the disabled wheelchair users, a constant and slow forward velocity $\nu$, along with an $\omega$ for adjusting the position of the chair on the corridor is sufficient for completing the task.

\begin{figure}[t]
      \centering
    \includegraphics[width=4cm]{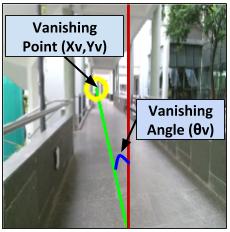}
      \caption[font=small]{Vanishing Point ($x_v$) and Vanishing Angle ($\theta_v$) features that are extracted by the TVS approach in \cite{following1}. These features are passed through a control law that computes a corrective velocity for autonomous corridor following. The red line is perpendicular to the corridor plane.
    }
    \label{fig: Vptheta}
    \vspace{-0.5cm}
   \end{figure}

In \cite{following1}, the authors describe a traditional approach for autonomous corridor following that makes use of vanishing point and vanishing line features to perform this task.
They use an automated feature extraction mechanism that adds constraints on the lines detected by the LSD algorithm \cite{lsd1,lsd2} from a captured image to obtain $x_{v}$ and $\theta_{v}$, the selected features for servoing. $x_v$ is the $x$ coordinate of the vanishing point, while $\theta_v$ is the angle that the vanishing line makes with the corridor plane. (Refer Figure \ref{fig: Vptheta})

In our approach, we use a human annotator to mark $x_v$ and $\theta_v$ features on an image. This ensures that the outcome of these features is reliable, as it is often the case that the automated feature extraction step fails to extract accurate features. This occurs mainly due to environmental noise in the captured image or sub-optimal feature extraction parameters which are difficult to tune.

For autonomous corridor following, the desired motion is achieved when the wheelchair is moving straight and is positioned at the center of the hallway. This occurs when $x_{v}$ lies at the center of the captured frame i.e the origin, and $\theta_{v}$ is perpendicular to the corridor plane in the image. The corresponding feature values of (0,0) are consequently chosen as the desired feature values.

In \cite{control}, a control law for servoing an image based path following system such as ours is formulated as:
\begin{equation*}\label{eq:Controllaw}
\omega = -J_{w}^{+}(\lambda e + J_{v}\nu^{*}) \tag{1}
\end{equation*}
The $\omega$ value here represents the ground truth angular velocity that we require to train our network. The Jacobians $J_{w}$ and $J_{v}$ are defined in \cite{following1} as follows:
\begin{equation}
\label{eq:Jw}
J_w = 
\begin{bmatrix}
    1 + x_{v}^{2} \\
    -\lambda_{\theta_v}lc + \lambda_{\theta_v} w \rho + \rho s  \tag{2}
\end{bmatrix}
\end{equation}
\begin{equation}
\label{eq:Jv}
J_\nu = 
\begin{bmatrix}
    0 \\
    -\lambda_{\theta_v}\rho           \tag{3}
\end{bmatrix}
\end{equation}
Here $x_{v}$ and $\theta_v$ are the selected features, $c = \cos{\theta_v}$, $s = \sin{\theta_v}$, $\rho = x_v\cos{\theta_v} \: + \: y_v\sin{\theta_v}$ and $\lambda_{\theta_v} = \cos{\theta_m}/h$. The $h$, $w$ and $l$ values represent the position of the camera on the wheelchair, which in our case is set to $h = 0.5$m, $w = 0$m, and $l = 0$m.
$\lambda$ and $\nu^{*}$ are gain and translational velocity constants that are tuned to $10^{2}$ and $0.2 m/s$ for our task.
The error $e$ is defined as the difference between the extracted features and the desired feature values.
\begin{equation}\label{eq:Error}
        e = (x_{v},\theta_{v}) - (0,0) \tag{4}
\end{equation}
The vanishing point coordinates are measured in meters and the slope of the vanishing lines are taken in radians.

The $\omega$ obtained here is the ground truth value used for training our convolutional neural network model.

\begin{figure*}[t]
      \centering
    \includegraphics[width=\linewidth]{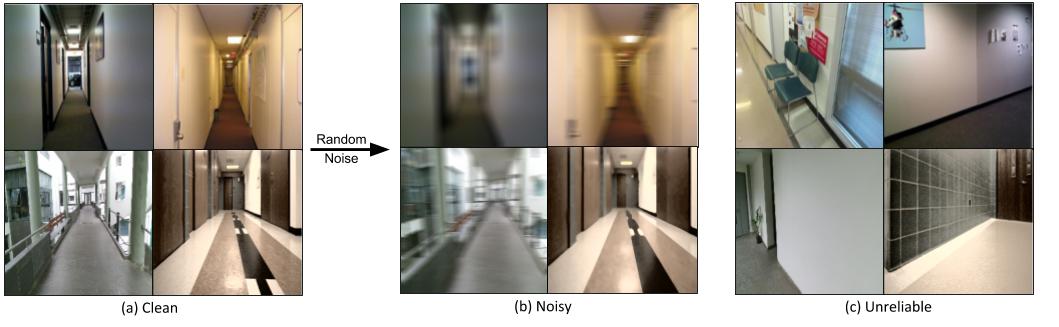}
      \caption[font=small]{Samples of images present in the dataset. The diversity in corridor environments can be observed in (a). Four different types of noises are artificially mixed with these images to obtain the noisy images shown in (b). Images where the required vanishing point features for computing the ground truth $\omega$ could not be extracted are shown in (c). These have been discarded from training.}
    \label{fig: Dataset}
    \vspace{-0.5cm}
  \end{figure*}
  
\begin{figure}[b]
    \vspace{-0.3cm}
      \centering
    \includegraphics[width=\linewidth]{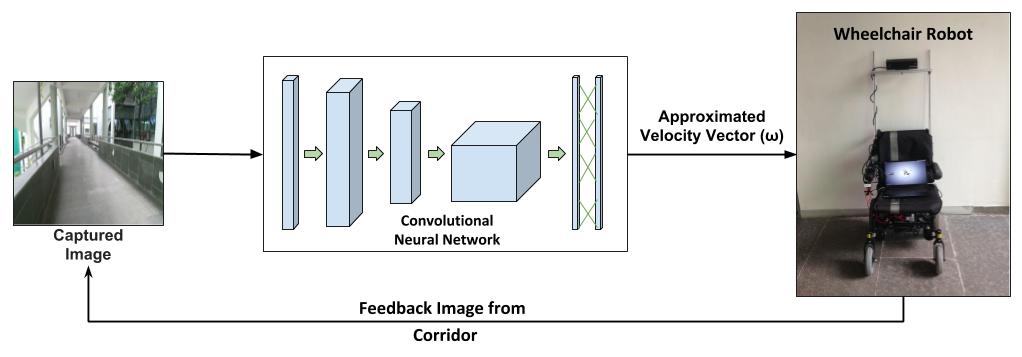}
      \caption[font=small]{Overview of our CNN Approach. An image captured from the corridor environment and passed through the CNN. The approximated velocity vector $\omega$ is then used to perform servoing on the wheelchair.}
    \label{fig: CNN Approach}
  \end{figure}
  

\subsection{The Training Dataset}
\label{Training_Dataset}
  
For training our network, we gather suitable corridor images from various open access sources \cite{dataset1,dataset2,dataset3,dataset4,dataset5}. We also create our own dataset of corridor images belonging to our institute. The accumulated set consists of 3563 images in total.
A small subset of 403 images belonging to this set have been discarded from training as their ground truths could not be estimated. This is because the vanishing point feature $x_{v}$ in these cases lies outside the frame of the image and cannot be extracted reliably. These images are deemed unreliable and Figure \ref{fig: Dataset}(c) shows examples of such cases.
The remaining samples compose a clean set of images.

We add 4 different types of artificial noise (Mild and Strong Gaussian Blur, Motion Blur and JPEG Compression) randomly to the entire clean set and obtain a separate noisy set of images. The final dataset is a combination of the clean and noisy sets and contains 6320 images. The ground truth values for samples in the noisy set are identical to their counterparts in the clean set. Fig \ref{fig: Dataset}(a) and \ref{fig: Dataset}(b) show examples of clean images their noisy counterparts.

Adding noisy images to our dataset serves a dual purpose of increasing the data used for training our model as well as helping the neural network generalize better to noisy data. 
The train-test split on the final dataset is 90-10\% and 10\% of the train set generated is used for validation. 
As the test set is randomly sampled from the final dataset, and contains a mixture of clean and noisy images.

\subsection{Designing and Training our CNN}
\label{sec:3}

Figure \ref{fig: CNN Approach} provides an overview of our visual servoing approach.
We employ a technique called transfer learning \cite{Transfer1,Transfer2} for training our model. In the most literal sense, transfer learning refers to "transferring" knowledge obtained from training one model to another model that performs a task of a different nature. This is especially useful in cases such as ours where the dataset size is small and there is insufficient training material for neural network to converge.

We exploit this technique and fine-tune a ResNet-18 architecture \cite{resnet} pre-trained on ImageNet for our task.
This setup was chosen considering ResNet-18's exceptional performance on ImageNet despite having a comparatively small model size. The model was pre-trained on ImageNet with an input size of $224$x$224$, and an output size of 1000 classes. All images in our dataset have accordingly been re-sized to $224$x$224$ to make sense of the pre-trained weights. We replace the final layer of this pre-trained model with a 1-dimensional output that represents the required angular velocity $\omega$ for our servoing task. We perform regression using this setup.

A Mean Squared Error (MSE) loss function determines the gradients for backpropagation for each iteration. In our case, this can be written as,
 \begin{equation}\label{eq:MSEloss}
     loss = \frac{1}{n}\sum_{i=0}^{n}(\hat{\omega}-\omega)^{2}                        \tag{5}
 \end{equation}
Here, $n$ is the batch size during training which has been set to 8.
$\hat{\omega}$ is the predicted angular velocity after a forward pass through the network and $\omega$ is the target angular velocity for that sample.

We train the network on an Nvidia 1080 Ti having 12GB of GPU memory and 64GB of RAM. It takes around 30 minutes for running 40 train-validation epochs. We employ a Stochastic Gradient Descent scheme with a weight decay of 0.005 and momentum of 0.9. The learning rate is set to 0.005. 10-fold cross validation was performed to understand the variance in training data and accordingly tune these network hyperparameters.

\subsection{Network Evaluation Metric}
\label{rsquare_description}
The $R^{2}$ value or coefficient of determination has been used as an evaluation metric for assessing the performance of the neural network on the task of regression. It can be defined in our case as:
\begin{equation}
    \label{rsquare}
    R^{2}(\omega,\hat{\omega}) = 1  - \sum_{i}^{n-1}\dfrac{(\omega_{i}-\hat{\omega})^{2}}{(\omega_{i}-\bar{\omega})^{2}}  \tag{6}
\end{equation}
Here, $\omega$ represents the true value i.e, the target distribution, $\hat{\omega}$ represents the predicted distribution, and $\bar{\omega}$ is the mean of the target distribution.
$n$ here represents the number of samples taken from the distribution, which in our case is the number of images in the test set.

The $R^{2}$ value ranges from $-{\infty}$ to 1. A positive value closer to 0 indicates that the model is unable to explain the variability of the data, while a value closer to 1 shows that the output corresponds well with the target distribution.

\section{Robustness Analysis of CNN-based Corridor following Scheme}
In this section, we discuss two methods for evaluating the performance of our CNN approach in cases when the TVS-based approaches like the one based on vanishing feature approach fails to perform well.
\label{Analysis}

\subsection{Comparing Deep and Vanishing Features}
\label{DeepvsClass}
When the vanishing feature method fails to produce a good $\omega$, it is often due to feature extraction going wrong. Our CNN approach is independent of this explicit feature extraction step and gives an approximation for $\omega$ even in these fail cases. In order to better understand the outcome of our CNN approach, we try to estimate the deep feature $x_v$ from the $\omega$ obtained by the CNN, and compare its performance against the vanishing point feature $x_v$ obtained by the vanishing point approach on the ground truth.

Recall the control law from \cite{control} presented in section \ref{sec:2}.
\begin{equation*}\label{eq:omegafirst}
\omega = -J_{w}^{+}(\lambda e + J_{v}\nu^{*})  \tag{7}
\end{equation*}
Here, the Jacobians $J_v$, $J_w$ and error $e$ are described in equations \ref{eq:Jw}, \ref{eq:Jv}, and \ref{eq:Error} respectively. After substituting these values and other constants mentioned in section \ref{sec:2}, we obtain:
\begin{multline*}\label{eq:omegasecond}
\omega = \begin{bmatrix}
        -1-x_v^2 & -\rho\sin{\theta_v}
        \end{bmatrix}
        \begin{bmatrix}
        \lambda x_v \\
        \lambda \theta_v
        \end{bmatrix}
        + \\
        \begin{bmatrix}
        -1-x_v^2 & -\rho\sin{\theta_v}
        \end{bmatrix}
        \begin{bmatrix}
        0 \\
        -\rho\cos{\theta_v}\nu^*/h
        \end{bmatrix}
\end{multline*}
This equation can be reduced to the following:
\begin{equation}\label{eq:omegaexpanded}
    \omega = -\lambda x_v -\lambda x_v^3 \: - \: \rho\lambda\theta_v\sin{\theta_v} + \frac{\rho^2}{h}\sin{\theta_v}\cos{\theta_v}\nu^{*}                \tag{8} 
\end{equation}
   \begin{table*}[t!]
    \centering
    \small
    \setlength\tabcolsep{1pt}
    \begin{tabular}{|p{0.14\linewidth}|c|c|c|c|c|}
        \hline
        \vspace{-1cm} \hspace{1.2cm} 1  &
        \includegraphics[width=0.166\linewidth]{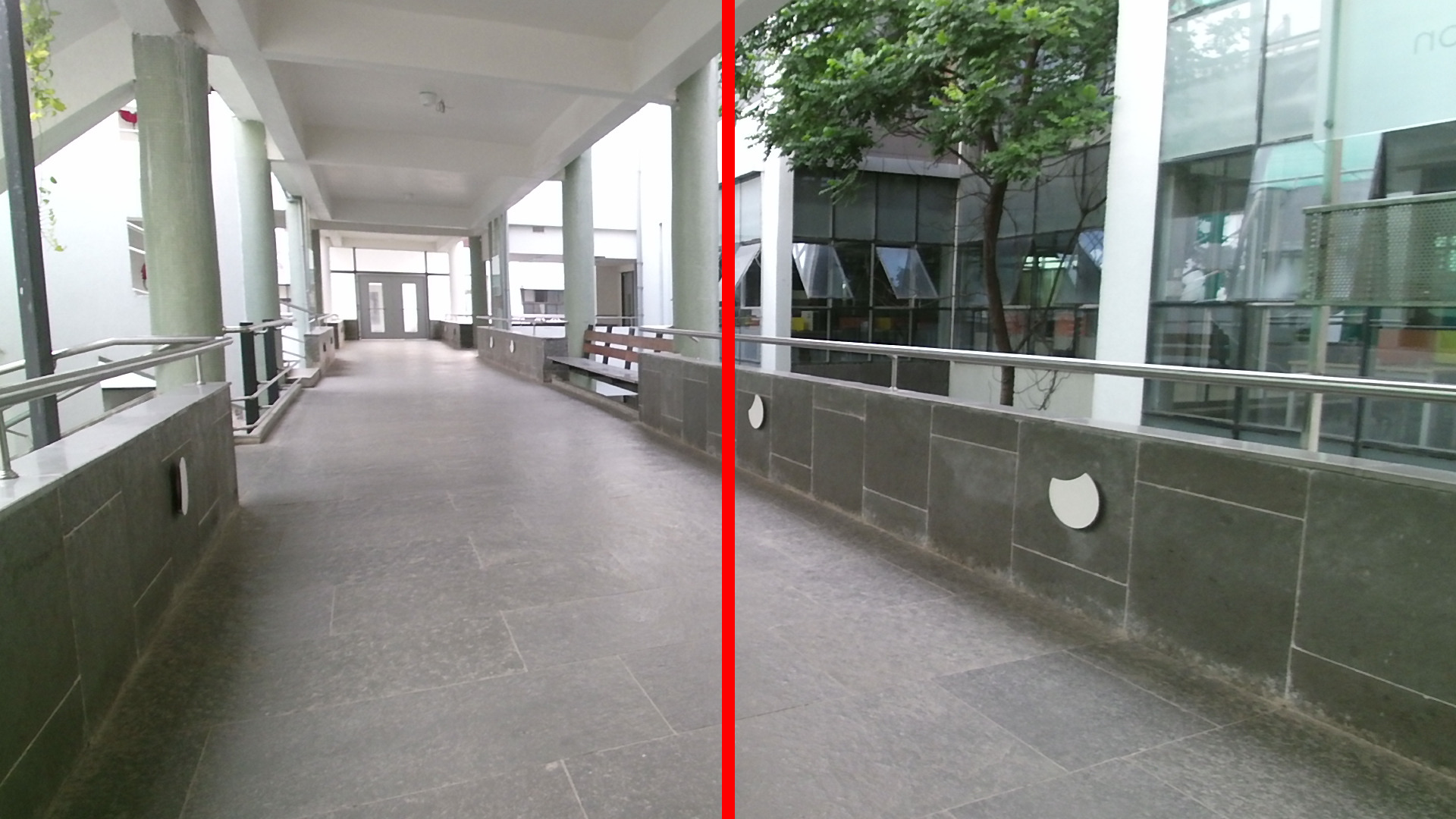} & 
        \includegraphics[width=0.166\linewidth]{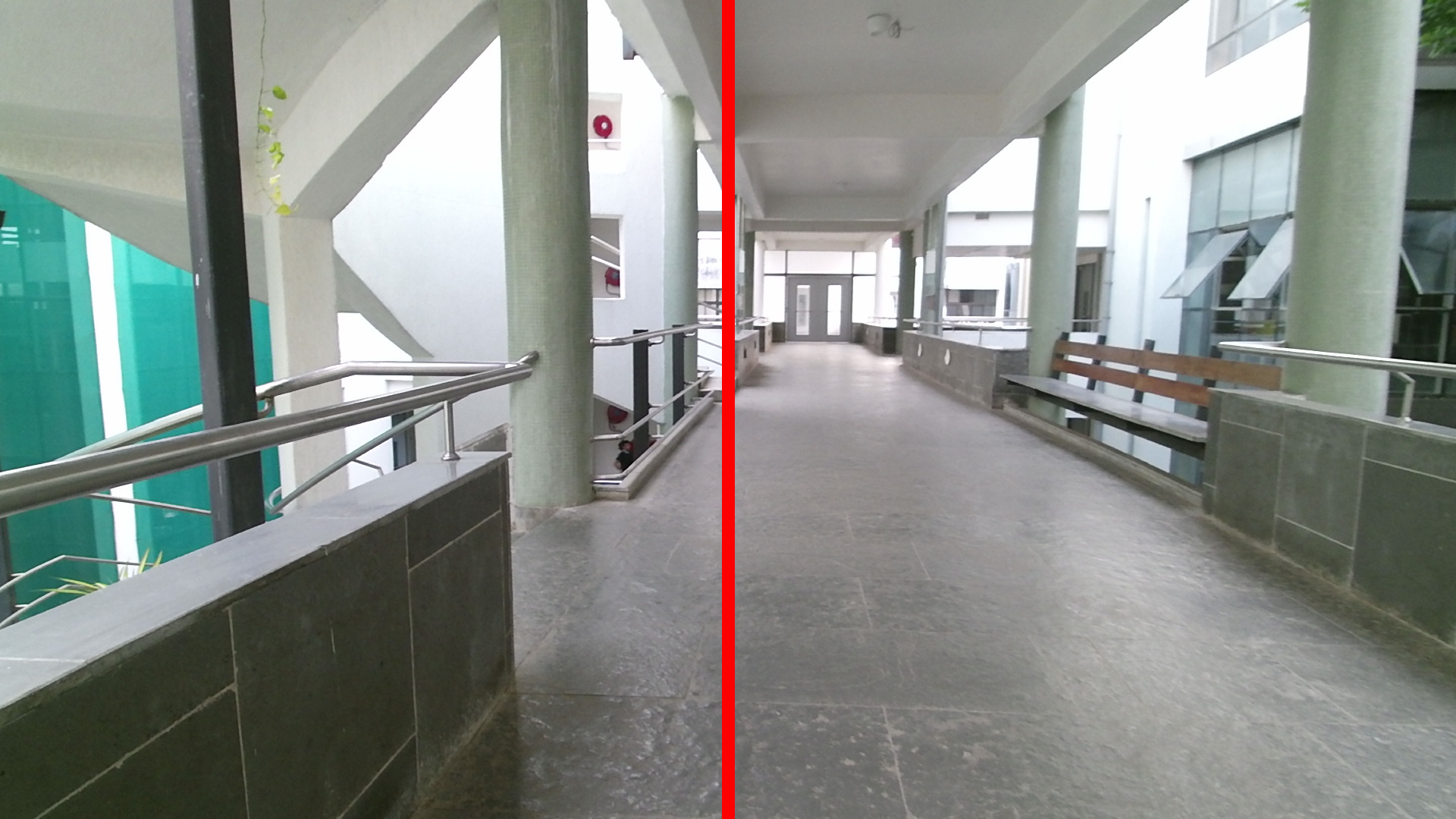} & 
        \includegraphics[width=0.166\linewidth]{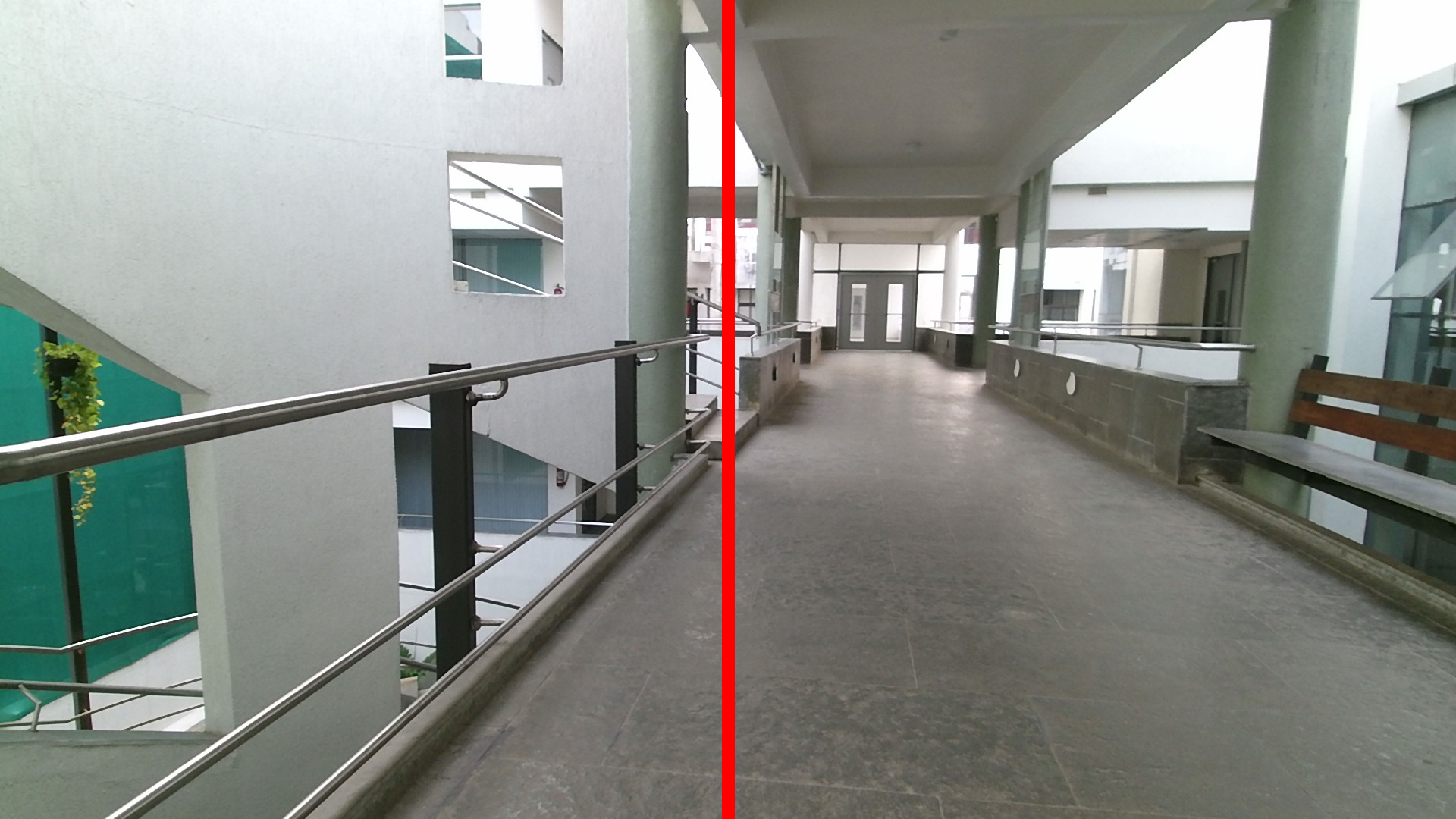} & 
        \includegraphics[width=0.166\linewidth]{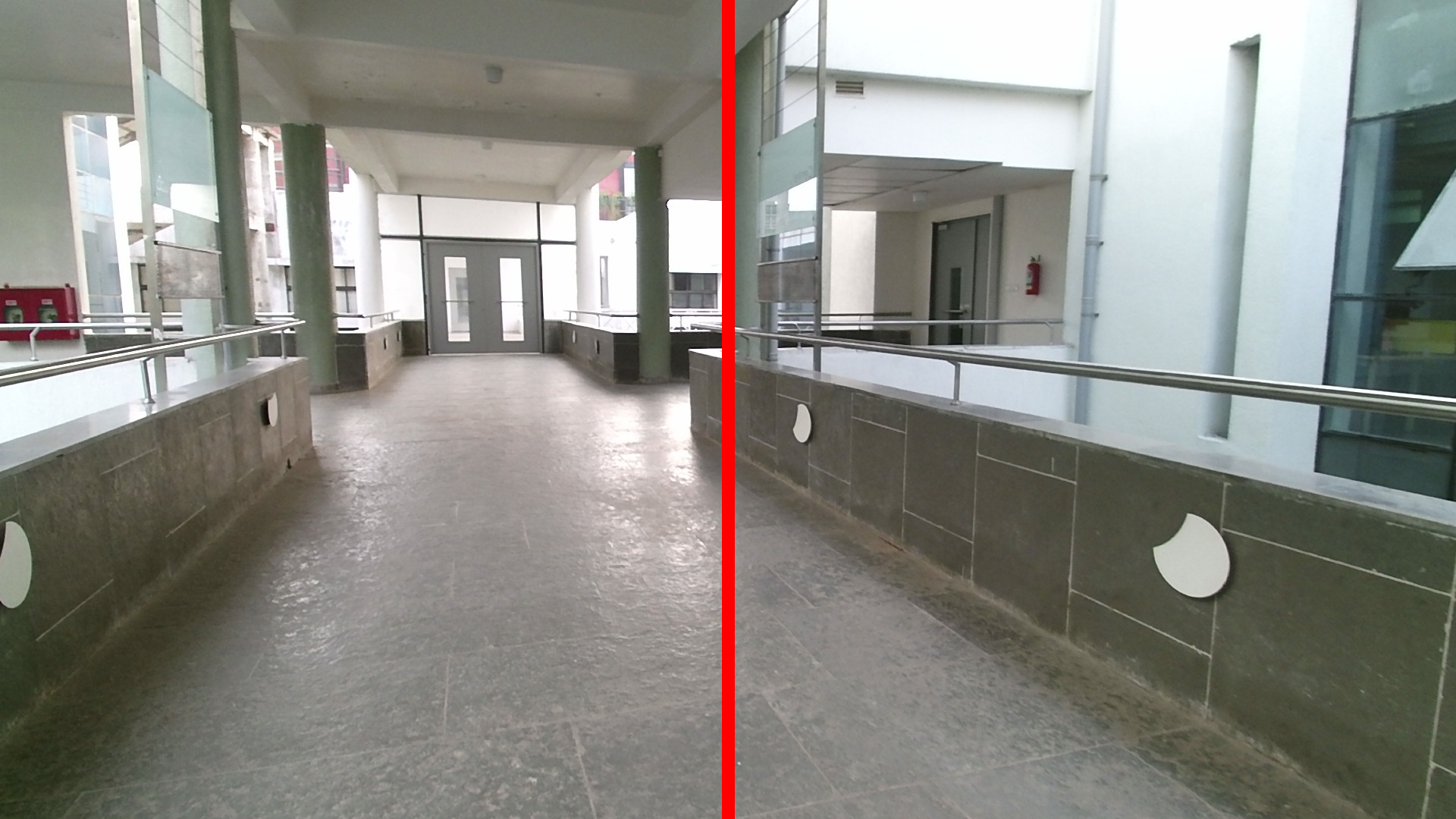} &
        \includegraphics[width=0.166\linewidth]{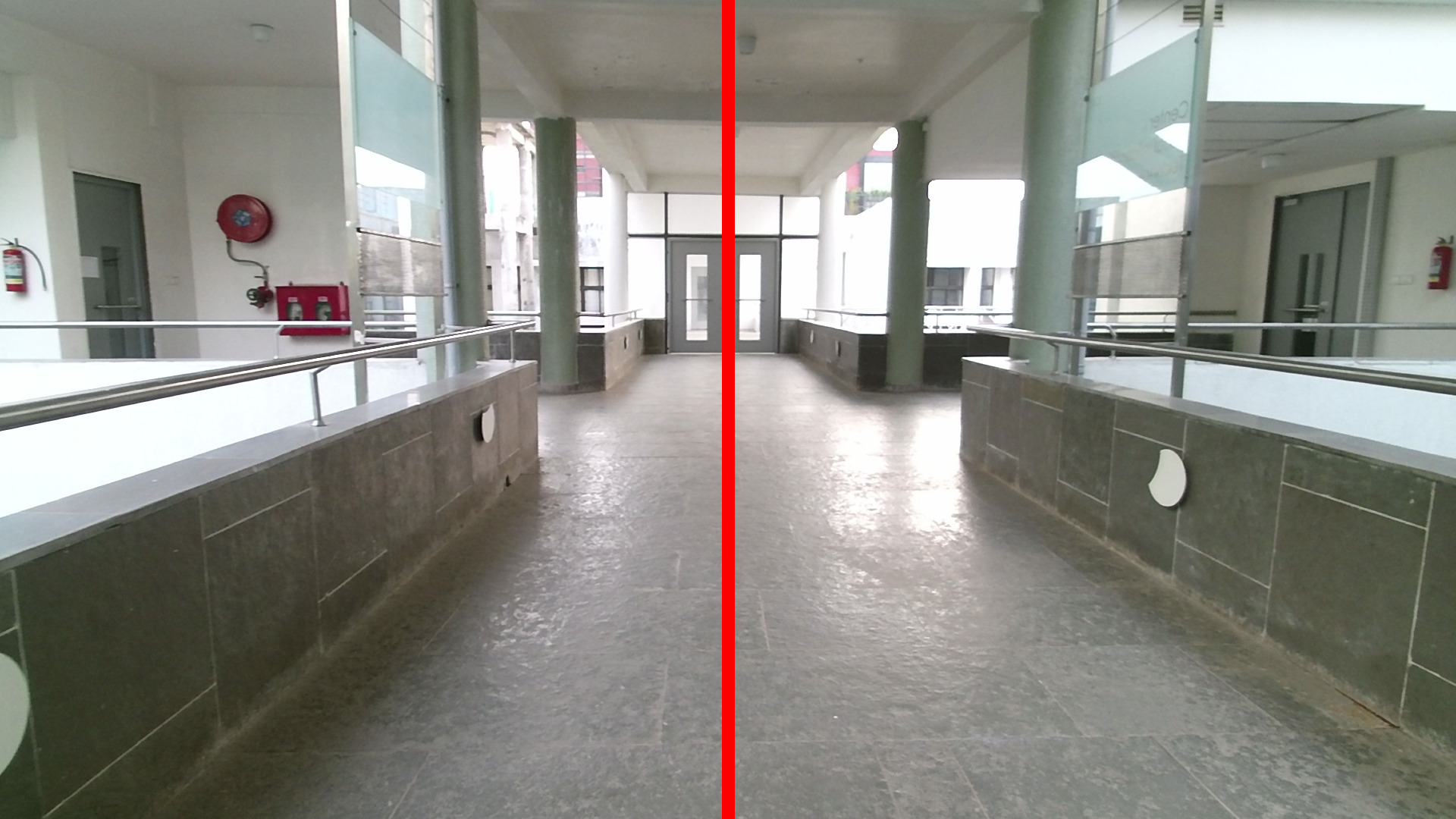} \\
        \hline
        CNN &        -1.067 & 0.289 & 0.500 & -0.622 & 0.059\\
        TVS-Based & -1.045 & 0.263 & 0.438 & -0.689 & -0.025\\
        Ground Truth &        -0.978 & 0.275 & 0.437 & -0.659 & -0.017 \\
        \hline
        \vspace{-1cm} \hspace{1.2cm} 2  &
        \includegraphics[width=0.166\linewidth]{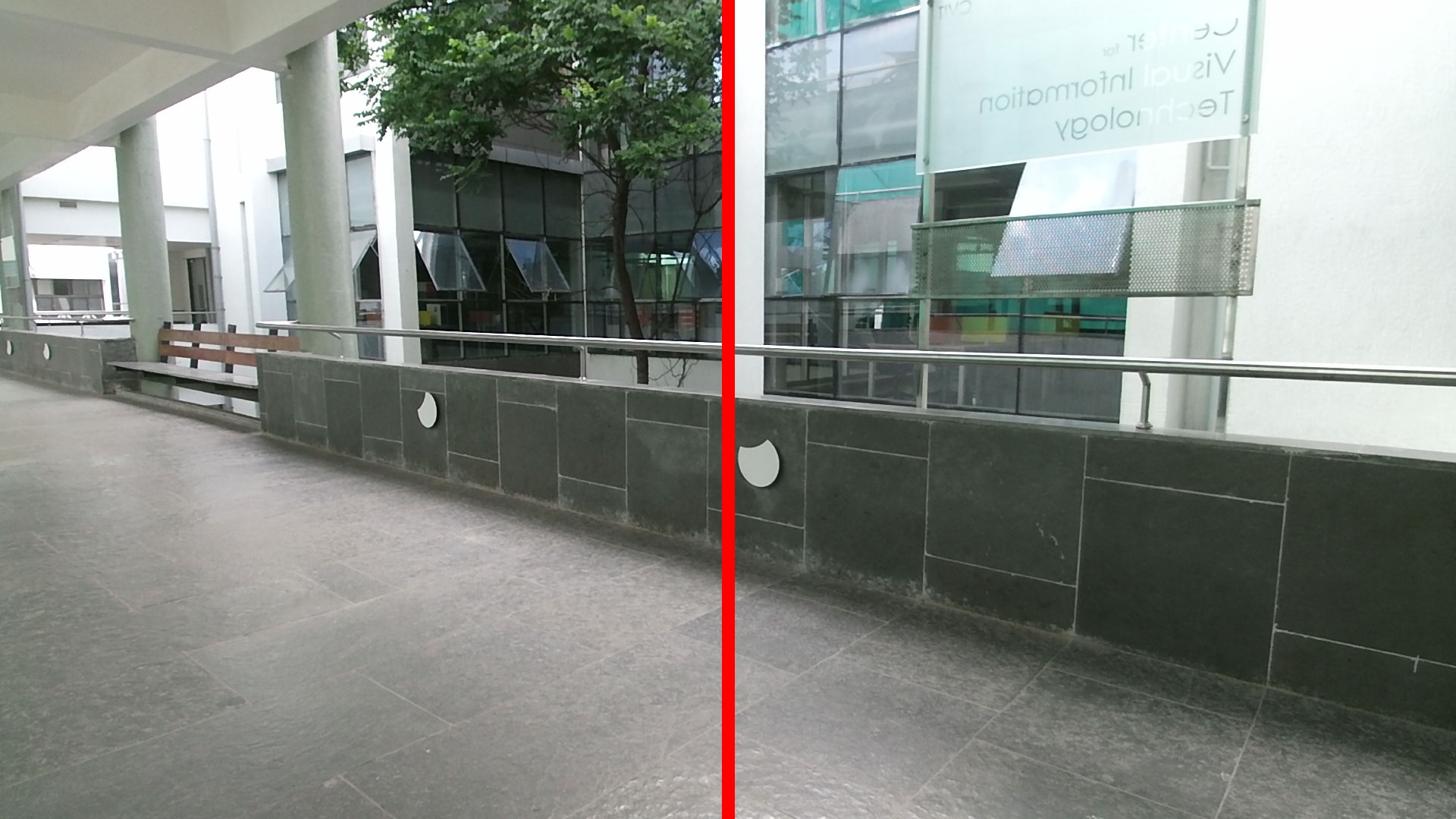} & 
        \includegraphics[width=0.166\linewidth]{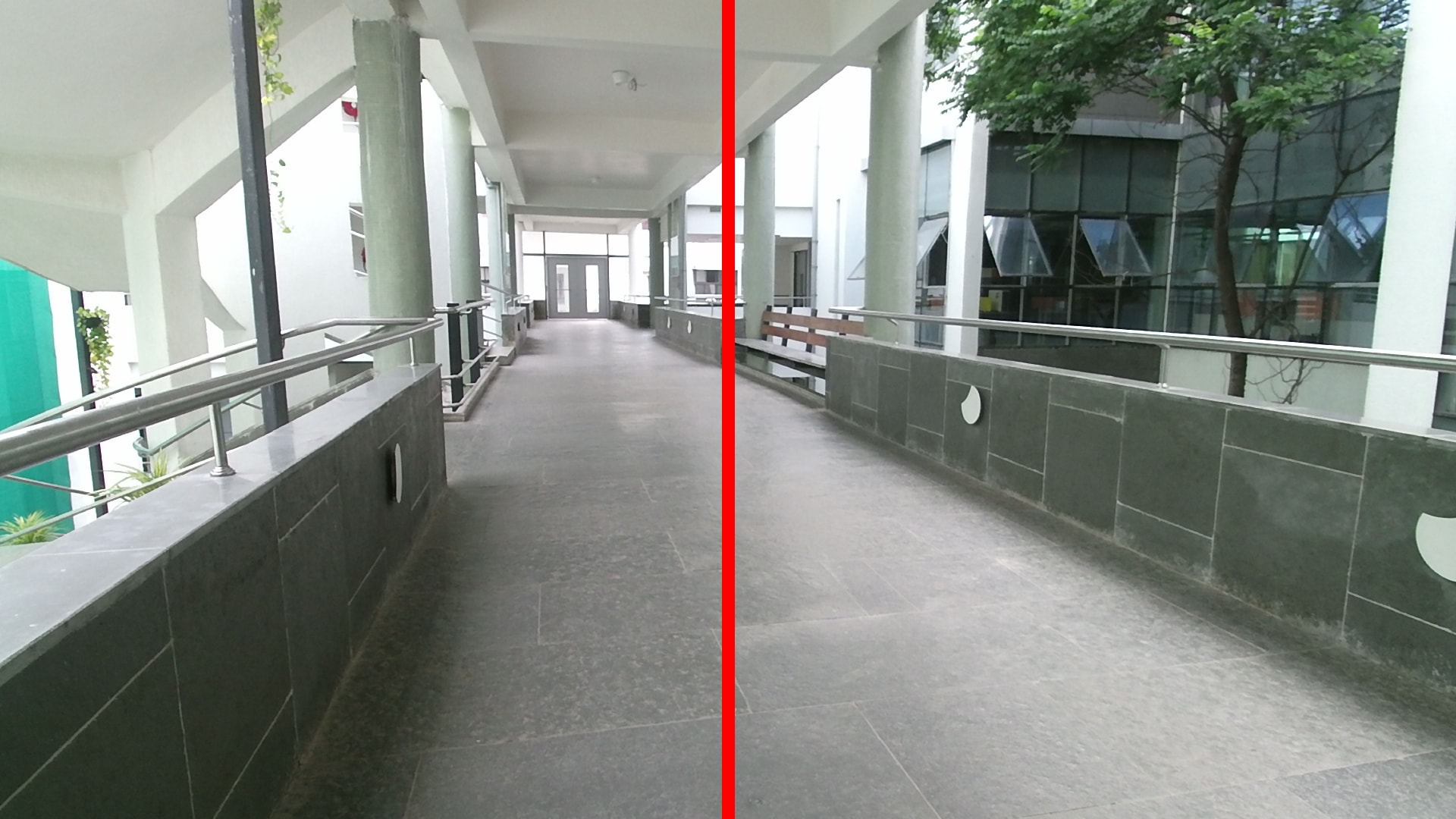} & 
        \includegraphics[width=0.166\linewidth]{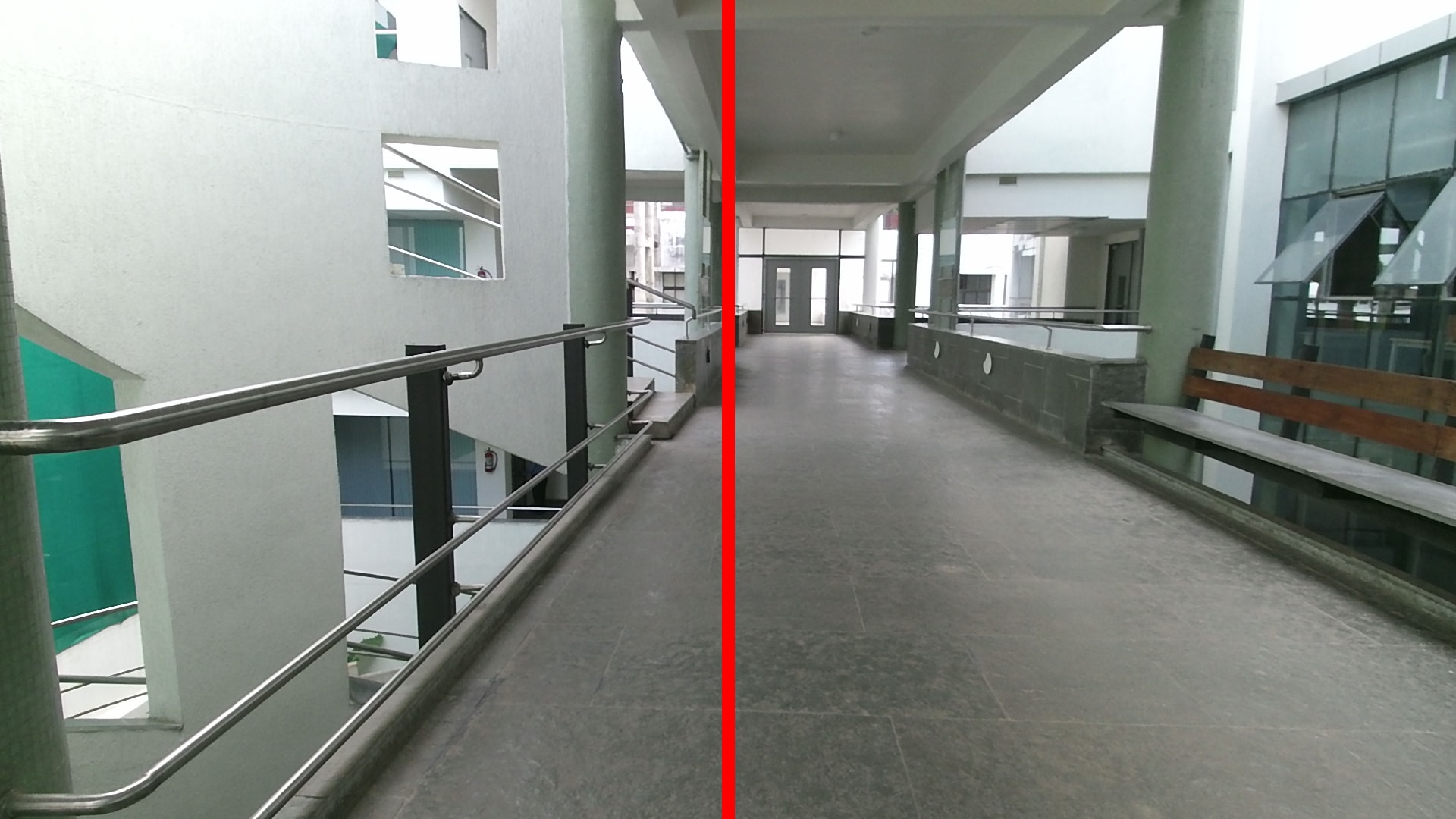} & 
        \includegraphics[width=0.166\linewidth]{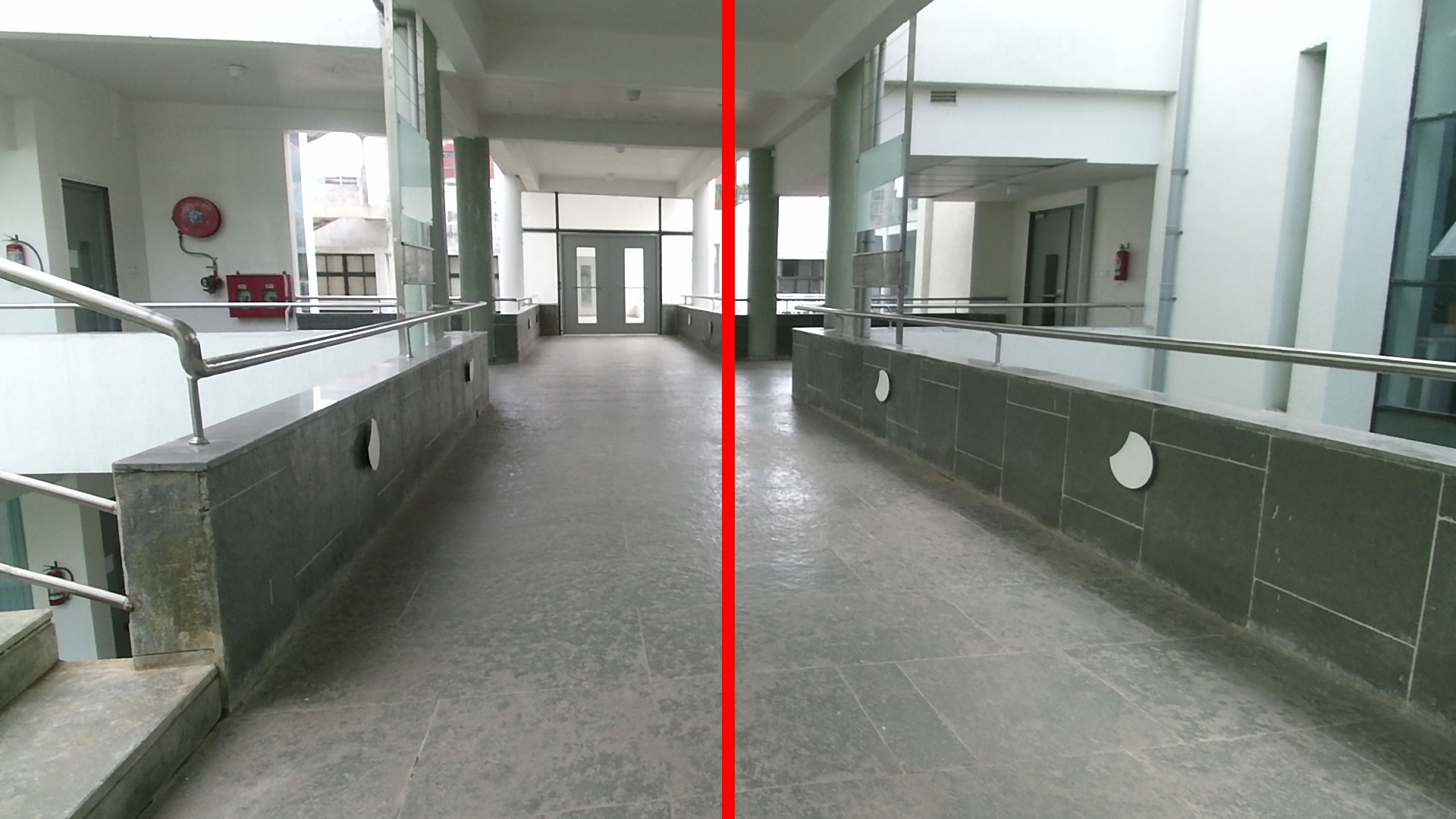} &
        \includegraphics[width=0.166\linewidth]{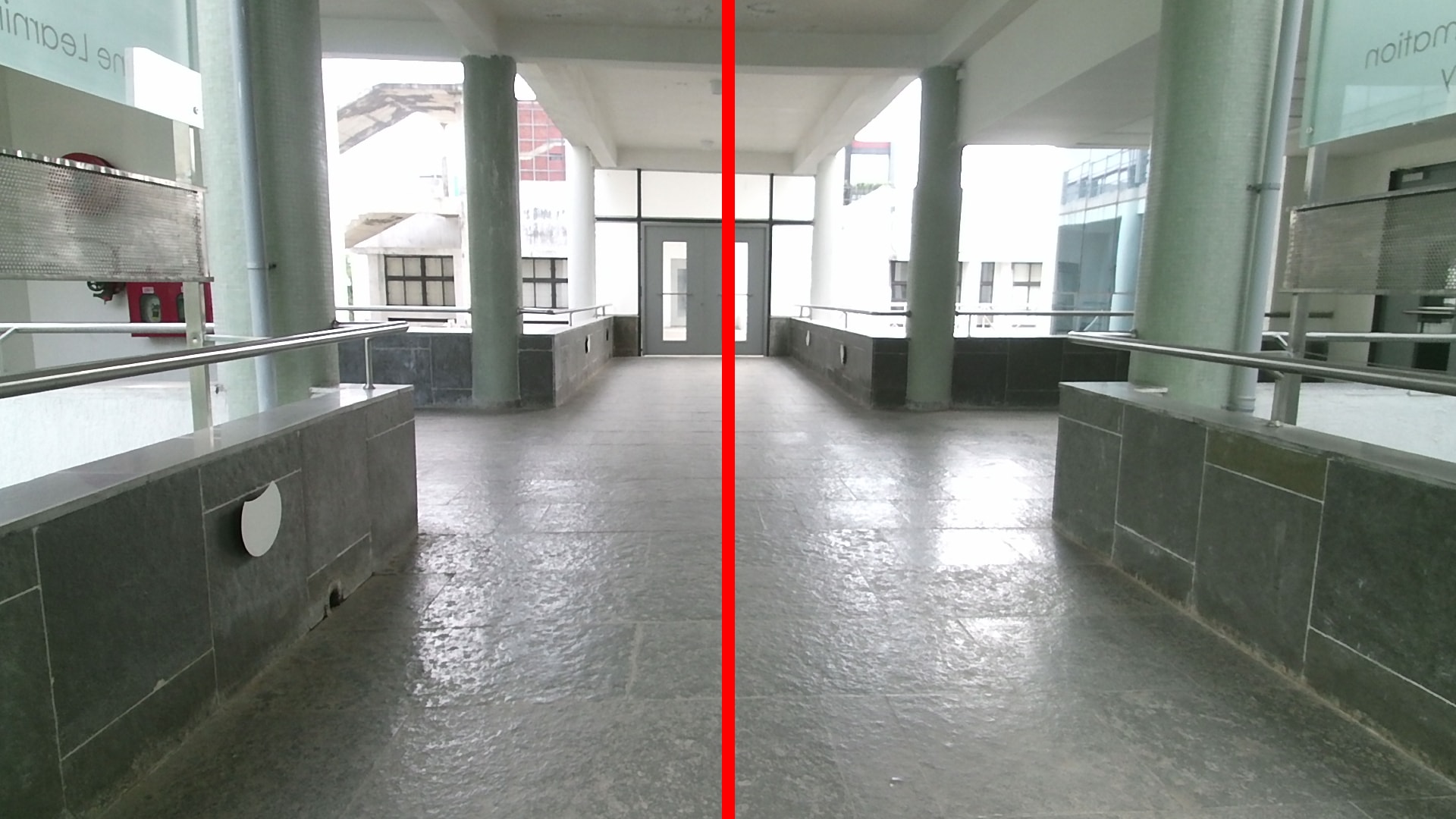} \\
        \hline
        CNN &     -1.472 & -0.450 & 0.117 & -0.346 & -0.027\\
        TVS-Based & NIL & -0.446 & 0.219 & -0.389 & -0.014\\
        Ground Truth &       NIL & -0.434 & 0.206 & -0.359 & -0.052 \\
        \hline
        \vspace{-1cm} \hspace{1.2cm} 3  &
        \includegraphics[width=0.166\linewidth]{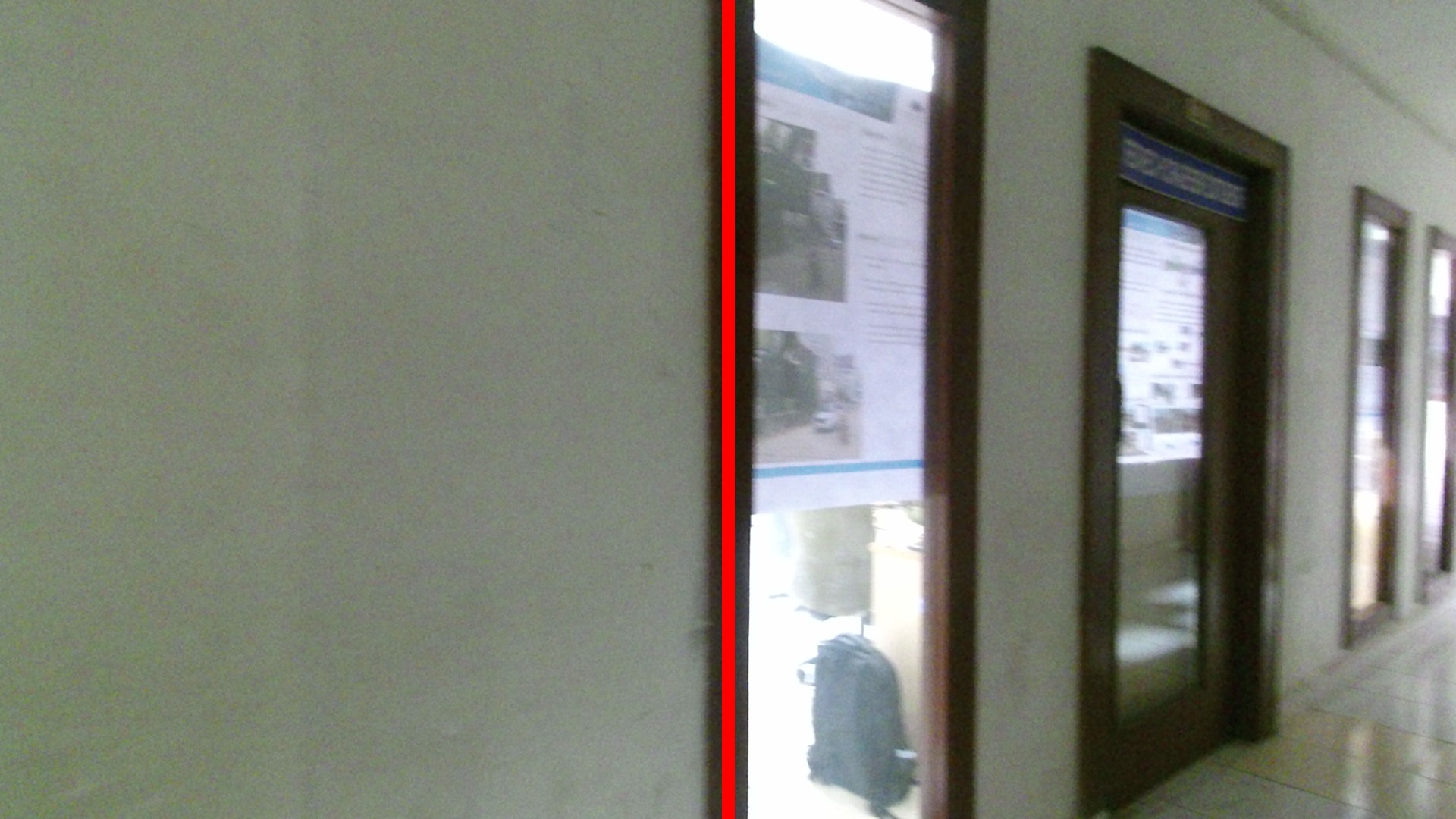} & 
        \includegraphics[width=0.166\linewidth]{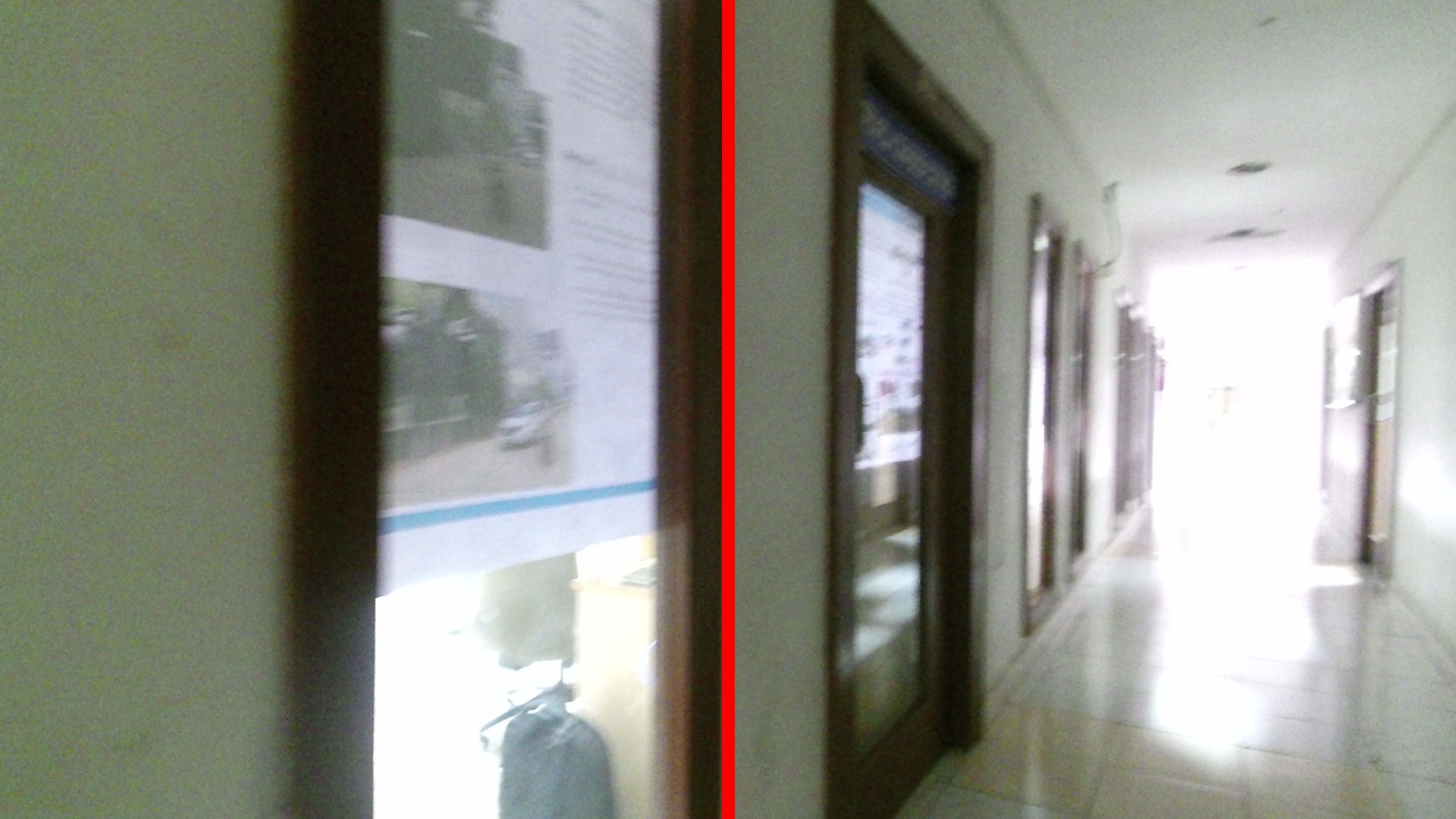} & 
        \includegraphics[width=0.166\linewidth]{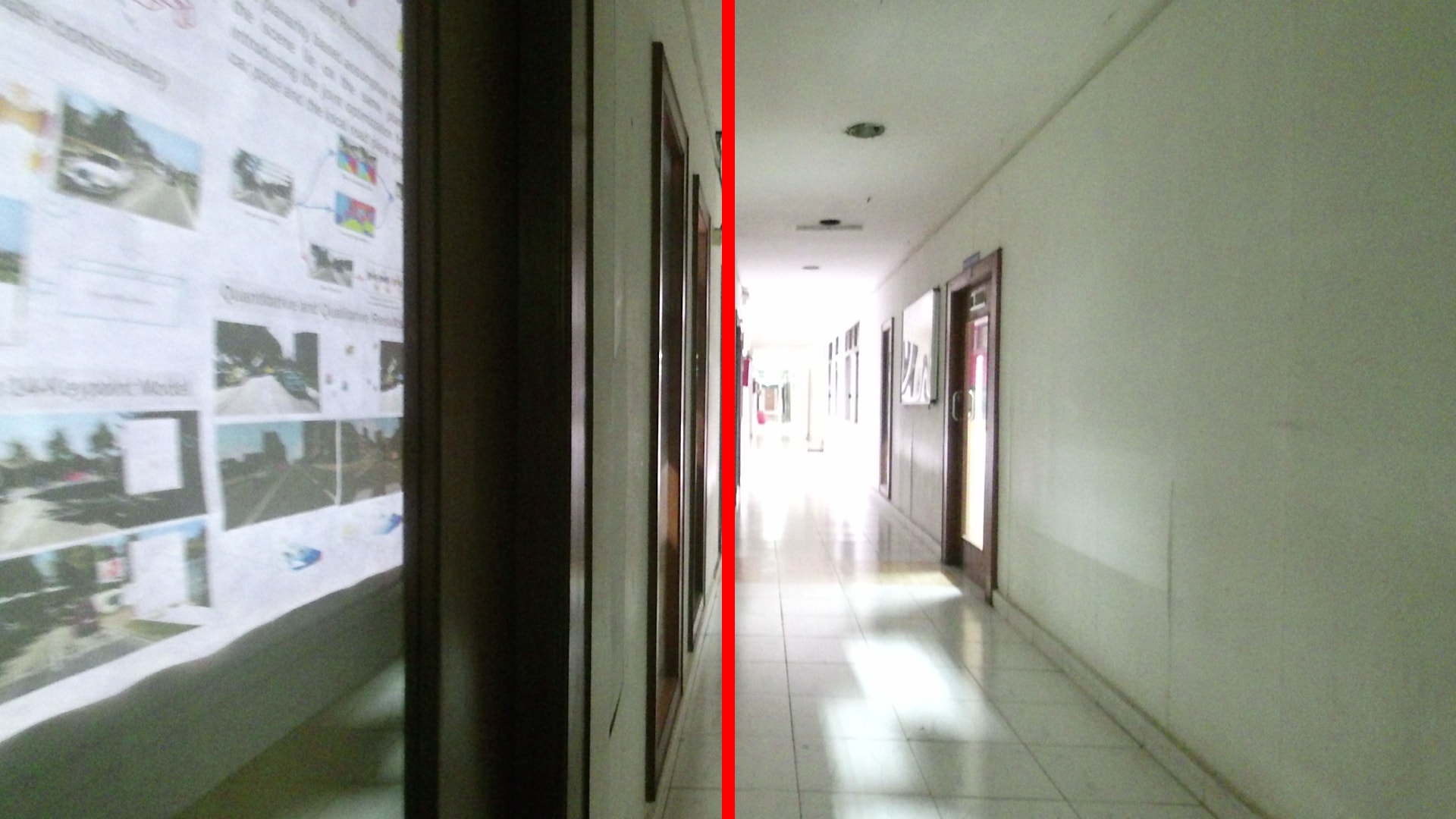} & 
        \includegraphics[width=0.166\linewidth]{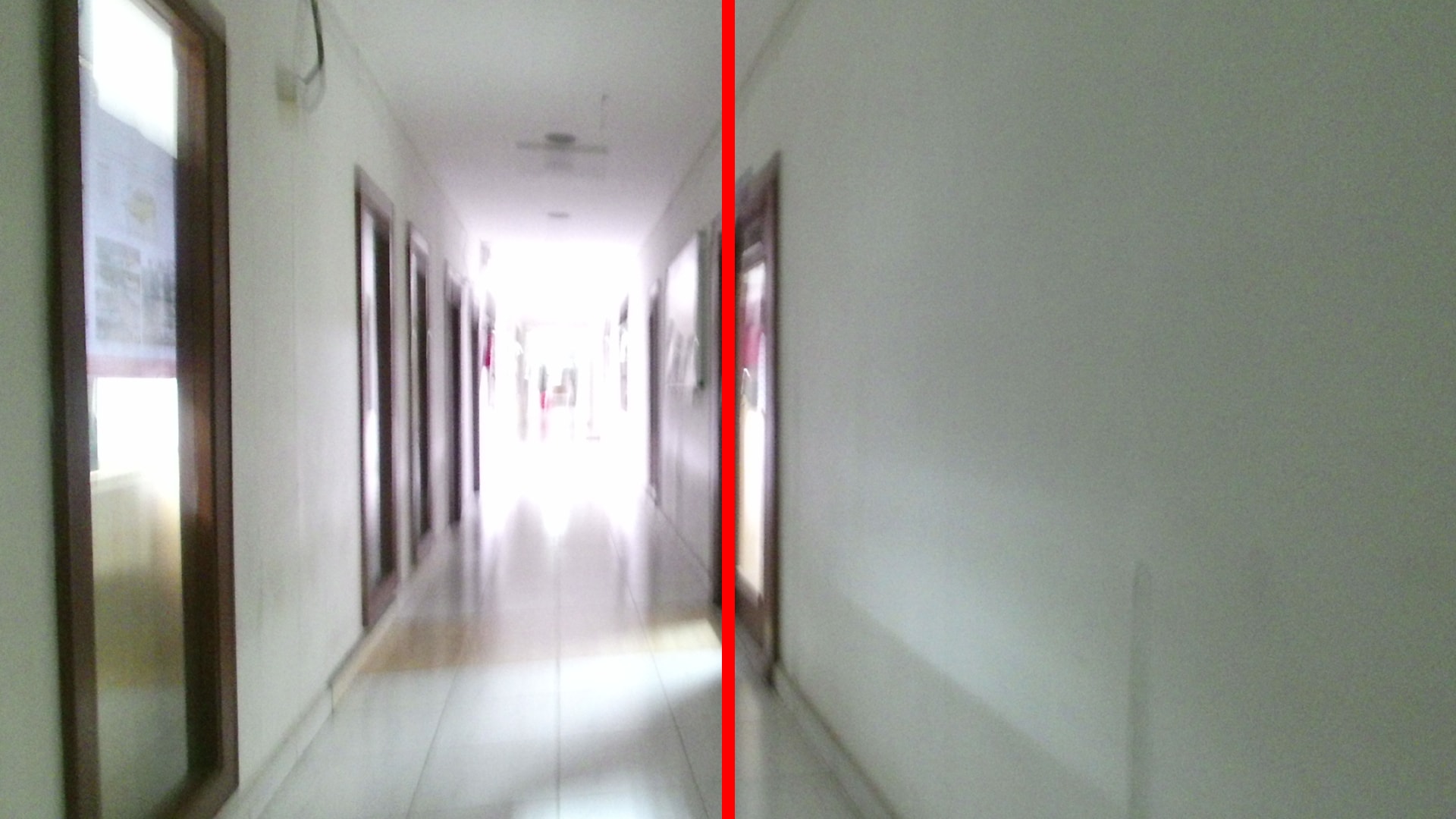} &
        \includegraphics[width=0.166\linewidth]{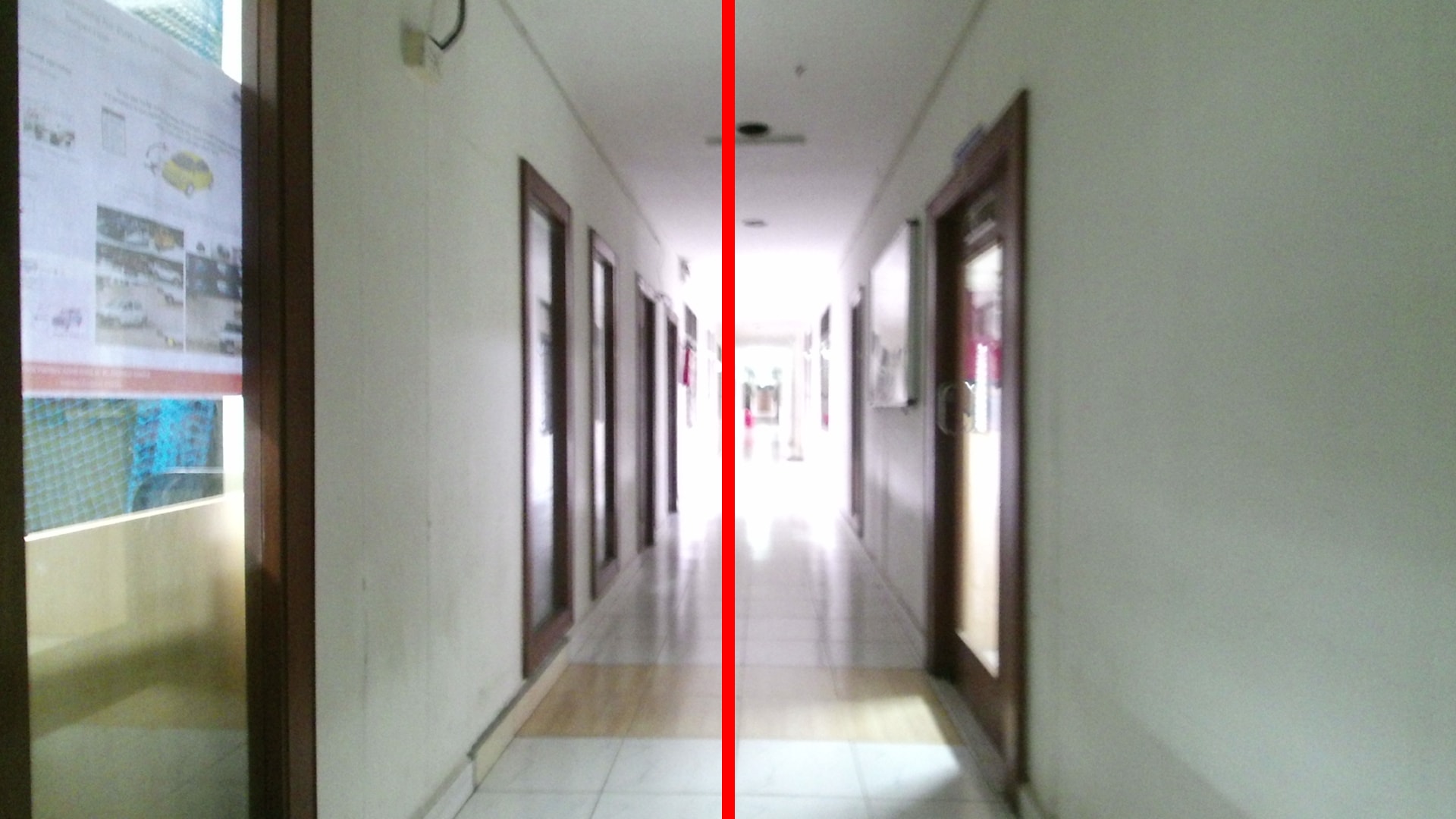} \\
        \hline
        CNN &       0.360 & 0.785 & 0.148 & -0.252 & -0.042\\
        TVS-Based & 2.687 & 0.227 & 0.146 & -2.397 & 0.052\\
        Ground Truth &        NIL & 1.376 & 0.109 & -0.445  & 0.03    \\
        \hline
    \end{tabular}
    \caption[name=Figure]{\textbf{Practical Results}: Image sequences captured from different corridors locations at our institute. The red line represents the ideal vanishing line when the wheelchair is at the center of the corridor and is used as a directional reference. The values are the estimated angular velocities for our CNN approach, TVS-Based vanishing feature approach, and a human annotated Ground Truth (GT). A positive $\omega$ value represents clockwise motion, and a negative value represents anti-clockwise motion.
    }
    \label{tbl:table_of_figures}
    \vspace{-0.5cm}
\end{table*}
Now, as we have only one equation with three variables, given $\omega$ we cannot find a closed form solution for $x_v$, $y_v$ and $\theta_v$. We can however limit the range of these values for our specific task to obtain a solution for a deep feature representing the vanishing point coordinate $x_v$ from $\omega$. Note that we do this only for illustrating the performance of our network against the TVS approach.

As the $x_v$ and $y_v$ values are represented in metres in the image plane, we can safely assume that their absolute values would be less than $1m$. Thus, the absolute value of $\rho$ i.e., $\sqrt{x_v^2 \;+\; y_v^2}$ also becomes less than $1$. Using this observation, we can neglect the last term in equation \ref{eq:omegaexpanded} as it is comparatively smaller to the other terms that contain $\lambda$, a large constant.

\begin{figure}[t!]
      \centering
    \includegraphics[width=\linewidth]{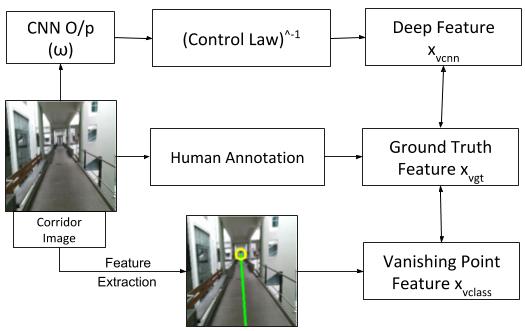}
      \caption[font=small]{Flowchart comparing Deep Features with TVS-based Vanishing Features. We perform a comparative analysis to understand why our CNN approach performs better in cases when feature extraction fails traditionally. This is achieved by extracting the $x_v$ feature using both approaches and comparing them against a common human annotated ground truth.}
    \label{fig: Comparison flowchart}
    \vspace{-0.5cm}
  \end{figure}

 From \cite{following1}, we can also conclude that $\theta_v \in (-\frac{\pi}{2},\frac{\pi}{2})$. However, in our experiments, we have observed that for most images in the dataset, $\theta_v \in (-\frac{\pi}{6},\frac{\pi}{6})$.
 Due to this reason, we chose to neglect the third term containing $\theta_v$ in equation \ref{eq:omegaexpanded} as it does not have a significant effect on the $\omega$ value in our case. This can also be experimentally observed by computing $\omega$ while changing the $\theta_v$ value.
 We can then solve the following equation for $x_v$:
\begin{equation}\label{eq:omegafinal}
    \omega = -\lambda x_v -\lambda x_v^3                \tag{9}
\end{equation}
As this is a third degree equation, with the discriminant $\Delta<0$, its real root $x_v$ can be expressed in terms of $\omega$ as:

\begin{align*}
    x_v = \sqrt[\scriptstyle 3]{\frac{-\omega}{2\lambda} + \sqrt{\Bigr(\frac{\omega}{2\lambda}\Bigr)^{2}+\frac{1}{27}}} + \sqrt[\scriptstyle 3]{\frac{-\omega}{2\lambda} - \sqrt{\Bigr(\frac{\omega}{2\lambda}\Bigr)^{2}+\frac{1}{27}}}
\end{align*}
$x_v$ here is a deep feature representing the vanishing point coordinate, that is obtained from the $\omega$ predicted by our CNN approach. This along with the traditional $x_v$ obtained from a TVS-based automated feature extraction mechanism is compared with the ground truth.
Figure \ref{fig: Comparison flowchart} shows a flowchart of our approach.

\subsection{Verifying Approximations for Unreliable Images}
\label{verification}
While training the model, we accounted for cases where the captured image is noisy, by adding noisy samples to the training data. However, there exist the unreliable images (Figure \ref{fig: Dataset}(c)) that were discarded from training the model as their ground truths could not be estimated using the traditional method. Our trained CNN model however can predict an approximation for $\omega$ for these images.

Just by looking at these unreliable images, a human can decipher if the wheelchair is meant to turn left or right to initiate the corridor following task. Also once a prediction is made, we know the direction of motion from the sign of the $\omega$ value.
We leverage upon these two pieces of detail for partially verifying the accuracy of the predicted outcome on unreliable images using human annotation.

For each unreliable image in the dataset, the following steps are taken by a human annotator:
\begin{itemize}
\item Pass the image through the trained network and obtain an approximation for $\omega$. Classify it as left or right based on the sign of the value predicted.
\item Show the same image to a human annotator equipped with a binary output console corresponding to the left or right direction.
\item Compare the human annotated output with the network output and update two score values representing accuracy and false positive accuracy.
\end{itemize}
Each annotator is shown the entire dataset 3 times, and an average of the scores obtained is chosen for evaluation.

The accuracy score tells us how well the network is able to predict the correct direction of motion from the image. It is the percentage of unreliable images that have had their outcome predicted in the right direction of motion.

\begin{equation}\label{eq:Accscore}
    Accuracy \: \: Score = \frac{n_i}{n} \%  \tag{11}
\end{equation}
Here, $n_i$ is the number of correctly predicted samples and $n$ is the total number of unreliable images.

The false positive score quantifies the severity of the network's bad performance on unreliable images.
It is the average of the absolute $\omega$ value on images where the wrong direction has been predicted.
\begin{equation}\label{eq:falsepositive}
    False \: \: Positive \: \: Score = \frac{\Sigma\omega_j}{n_j}     \tag{12}
\end{equation}
Here, $n_j$ is the total number of false positive samples and $\omega_j$ is the angular velocity predicted for these samples.

For autonomous corridor following in unreliable cases, a high accuracy score and low false positive score is desired.

\section{Experiments and Results}

\label{Results}
\subsection{Evaluation of Neural Network Performance}
We evaluate our trained model on 4 noisy test sets in addition to the original test set using the $R^{2}$ score described in section \ref{rsquare_description}. Each noisy set comprises of images having one specific type of artificial noise. 
Table \ref{rsquared_table} shows the percentage $R^{2}$ value for each test set. Here, a similar score in all the test sets shows that the performance of the CNN on the noisy sets is as good as its performance on the original, non-noisy set, thereby establishing robustness to noise.

\begin{table}[t]
\centering
\begin{tabular}{|c||c|}
\hline
Test Set Type & $R^{2}$ Value(\%) \\
\hline
Original (Clean) & 88.321\\
\hline
Motion Blur & 88.011\\
\hline
JPEG Compression & 88.572\\
\hline
Gaussian Blur & 88.340\\
\hline
\end{tabular}
\caption{\textbf{Comparison of $R^{2}$ Values on Test Sets}: Here, as the $R^{2}$ values are similar across all the test sets, we can safely conclude that the performance of the neural network on noisy images is on par with that of clean images.}
\label{rsquared_table}
\vspace{-0.5cm}
\end{table}
On the unreliable test set, following the human verification method described in the earlier section (\ref{verification}), we get an accuracy score of $78.75\%$ on predicting the right direction of motion on 403 unreliable images.
We get a false positive score of $0.180$ on the 88 images that were predicted wrong. This translates to $5.16\%$ of the highest $\omega$ value obtained in our tests which shows that even when the wrong direction is predicted, the magnitude of the velocity vector output remains relatively small. Although this is specific to our case, it is a significant result as the CNN performs well on these images where the traditional vanishing feature approach would fail entirely.
\subsection{Practical Implementation and Results}\label{sec:5}
We practically evaluate our method on an Intelligent Wheelchair Platform developed at IIIT, Hyderabad\footnote{\href{https://youtu.be/aRGVXq8cqDs}{https://youtu.be/aRGVXq8cqDs}}.
A Kinect v2 has been retrofitted onto this platform as a sensor for capturing images. All processing is done on board on a laptop having an Nvidia 1050 Ti with 4GB GPU memory and 8GB RAM. A Sabertooth motor controller attached to the wheelchair takes serial commands from the laptop and translates them into actuary signals that controls its motion.
The entire epoch time from capturing an image to actuation takes $\approx1.8$ seconds or $\approx0.6$Hz on this setup. Although slow for many real time systems, this control frequency is sufficient for our task as our final application is on an assistive wheelchair for the disabled.
Our translational velocity is set to $0.2$m/s, ensuring that the wheelchair is slow enough for sufficient coherency between subsequent frames.

We conduct autonomous corridor following experiments on different corridors across our institute, including environments that were previously unseen in the training dataset. In each experiment, the wheelchair is made to start at an arbitrary position at the beginning of the corridor making an arbitrary angle between $[\ang{0},\ang{90}]$ with the wall. The corridor following task is then carried out using the proposed CNN method and the images captured are stored along with their corresponding CNN $\omega$ values. We then use the traditional vanishing feature approach to estimate an $\omega$ on these stored images, and also a ground truth $\omega$ using human annotation (Refer \ref{Velocityestimateclassical}).

Table \ref{tbl:table_of_figures} has samples of image sequences captured during the experiment and their corresponding $\omega$ values for the CNN, vanishing feature approach and the ground truth.
There is a high correspondence between the values of the CNN and vanishing feature approach in sequence 1.
In sequence 2 however, when the wheelchair starts at a sharper angle with the corridor wall, an `unreliable image' is captured due to which the traditional $\omega$ does not get computed. A ground truth does not exist here either, as a human annotator cannot accurately mark features outside the image frame. The CNN approach here predicts a velocity in the anti-clockwise direction, which enables the wheelchair to initiate and follow through the servoing task.

Sequence 3 is taken from an environment outside the training dataset. Observe the erratic $\omega$ values that the traditional approach estimates due to bad feature extraction. This is primarily because we do not re-tune the traditional approach parameters for extracting features from this environment. The CNN on the other hand first predicts a small value for servoing on the unreliable image captured in the beginning. Once the corridor is fully visible, it predicts a better approximation closer to the GT value, and successfully completes the servoing task.

\begin{figure}[t]
      \centering
    \includegraphics[width=\linewidth]{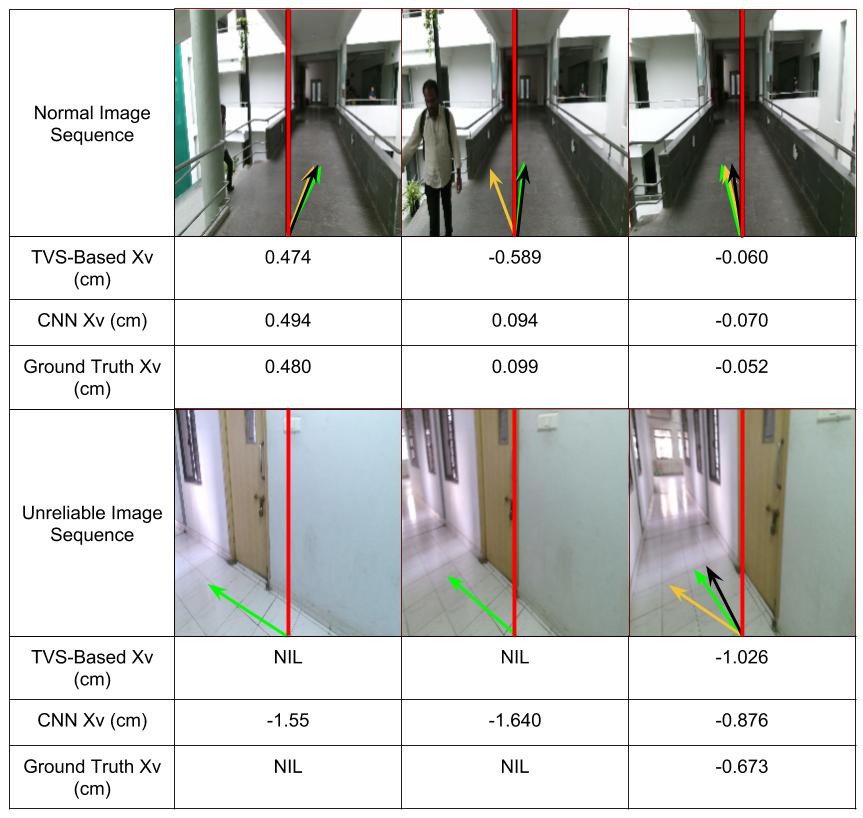}
      \caption[font=small]{Comparison of Deep vs Traditional Features. The yellow arrow represents the traditional approach, the green one represents the CNN approach, while the black one represents the ground truth. The red line at the center of the image is for directional reference. The angles that the arrows make with this line are directly proportional to $x_v$. In the normal sequence, observe the disruptive value of the traditional $x_v$ when a person (environmental noise) enters the frame. In the unreliable sequence, observe the inability of the TVS method to extract the $x_v$ feature in the first two images.
      }
    \vspace{-0.5cm}
    \label{fig: Comparison x_v}
\end{figure}
\subsection{Advantages of the CNN approach}
We use the method described in \ref{DeepvsClass} to extract $x_v$ from the CNN $\omega$, and compare it with the vanishing point outcome (Refer Figure \ref{fig: Comparison x_v}).
\begin{enumerate}

\item Robustness to Environmental Noise: 
As our CNN is trained offline on several images of various corridor environments (both noisy and non-noisy) it works well on different environments including ones that are dynamically changing. 

The normal image sequence in Figure \ref{fig: Comparison x_v} illustrates this with an example. In the second image, when a person enters the frame, the traditional approach fails to compute a correct $\omega$, as its feature extraction step that is dependant on a line detector fails. The $x_v$ obtained is thus not representative of the actual vanishing point. Our CNN approach on the other hand predicts a velocity $\omega$ in the correct direction, which is backed up by a good deep feature $x_v$ extracted from the image.

\item Approximations for Unreliable Images:
As mentioned earlier in Section \ref{sec:2}, using the traditional method for servoing fails on unreliable images. This is due to the feature extraction step failing by virtue of the required vanishing point feature $x_v$ not lying on the image frame. Here, even if $x_{v}$ is extracted as an extended coordinate, its value is cannot be verified. A very large $x_v$ can cause the control law parameters to ``explode'' leading to the calculation of an unstable $\omega$. This holds true especially in cases where the $x_{v}$ extracted tends to $\infty$. The control law reaches a mathematical singularity here.

Observe the unreliable image sequence in Figure \ref{fig: Comparison x_v}. Here, in the first two images, as a well defined vanishing point $x_v$ does not exist, the traditional method fails. However, our CNN estimates $x_v$ as a deep feature, that enables motion in the correct direction, due to which corridor following becomes feasible.
\end{enumerate}

\section{Conclusion and Future Work}
We have shown that our approach has an advantage where a velocity outcome is predicted regardless of the input image. This can also be a disadvantage in some cases, where the wheelchair needs to stop in order to complete the task. To overcome this, future work may include training the CNN with `end of the corridor' and `object of interest' cases where the wheelchair would have to stop and reconsider its position before moving.

There is also a disadvantage in terms of the time taken for gathering a dataset for the purpose of corridor following, which is not required in traditional schemes. We plan to release the dataset of corridor images along with their human annotated ground truths to alleviate this issue for other researchers.

In conclusion, we have presented an end to end CNN based approach for autonomous corridor navigation on a wheelchair. Our network is trained to predict a velocity signal for the servoing task from a captured image. In doing this, our method overcomes some key limitations of a traditional visual servoing based approach. We demonstrate these by performing a statistical and experimental validation of our approach against the traditional approach.




 \addtolength{\textheight}{-8cm}   
                                  
\printbibliography
\end{document}